# The Assumptions Behind Dempster's Rule

Nic Wilson
Department of Computer Science
Queen Mary and Westfield College
Mile End Rd., London E1 4NS, UK

## Abstract

This paper examines the concept of a combination rule for belief functions. It is shown that two fairly simple and apparently reasonable assumptions determine Dempster's rule, giving a new justification for it.

## 1 INTRODUCTION

Dempster's rule is the cornerstone of Dempster-Shafer Theory, the theory of uncertainty developed by Shafer [76a] from the work of Dempster [67]. The rule is used to combine the representations of a number of independent evidences, to achieve a combined measure of belief. For the theory to be able give meaningful conclusions, it is essential that Dempster's rule is convincingly justified. The rule and its justifications have been criticised from many angles, a common criticism being that it can be hard to know when evidences are independent, and indeed, what 'independence' means here.

In this paper an axiomatic approach to the combination of belief functions is taken. The concept of a combination rule is formulated precisely, and assumptions are made which determine a unique rule, Dempster's rule. A benefit of this approach is that it makes the independence or irrelevance assumptions explicit. Since the assumptions are arguably reasonable this gives a justification of the rule. This justification is quite different from previous justifications of the complete rule, though it is related to the justification in [Wilson, 89, 92c] of Dempster's rule for a collection of simple support functions.

In section 2, the mathematical framework is introduced; in section 3, the concept of a combination rule is defined; section 4 discusses Dempster's rule and some of the problems with previous justifications of the rule; section 5 defines Bayesian conditioning, used for representing one of the assumptions; the assumptions on rules of combination are defined and discussed in section 6, and the main result of the paper, that they determine Dempster's rule, is given.

## 2 SOURCE STRUCTURES AND BELIEF FUNCTIONS

In this section the basic concepts are introduced. The mathematical framework is essentially that of [Dempster, 67] with different notation (and minor differences) but some fundamental issues are considered in greater detail.

### 2.1 SOME BASIC CONCEPTS

We will be interested in sets of propositions and considering measures of belief over these.

**Definition: Frame**

A frame is defined to be a finite set[1].

Without loss of generality, it will be assumed that frames are subsets of the set of natural numbers,[2] $\mathbb{N}$.

The intended interpretation of a frame is a set of mutually exclusive and exhaustive propositions. Then the set of subsets of a frame $\Theta$, written as $2^\Theta$, is a boolean algebra of propositions.

---

[1] 'Frame' is an abbreviation for Shafer's term 'frame of discernment' [Shafer, 76a]; [Dempster 67, 68] and [Shafer, 79] allow frames to be infinite; however the results here only apply to finite frames of discernment.

[2] Actually any other infinite set would do; this is just to ensure that the collection of all multiple source structures (defined later) is a set.

**Definition: (Additive) Probability Function**

Let $\Omega$ be a frame. P is said to be a probability function over $\Omega$ if P is a function from $2^\Omega$ to $[0,1]$ such that (i) $P(\Omega) = 1$, and (ii) (additivity) for all $A, B \subseteq \Omega$ such that $A \cap B = \emptyset$, $P(A \cup B) = P(A) + P(B)$.

We are interested in the propositions in $2^\Theta$, for frame $\Theta$. Dempster, in his key paper [Dempster, 67] considers a situation where we have a probability function over a related frame $\Omega$ representing Bayesian beliefs.

**Definition: Source Structure**

A source structure[3] $S$ over frame $\Theta$ is a triple $(\Omega, P, I)$, where $\Omega$ and $\Theta$ are frames (known as the underlying frame and frame of interest respectively) P (known as the underlying probability function) is a probability function over $\Omega$, and *compatibility function I* (Dempster's multi-valued mapping) is a function from $\Omega$ to $2^\Theta$. Furthermore, for $\omega \in \Omega$, if $I(\omega) = \emptyset$, then $P(\omega) = 0$.

The interpretation of $S$ is as follows. The set of propositions we are interested in is $2^\Theta$, but we have no uncertain information directly about $\Theta$. Instead we have a subjective additive measure of belief P over $\Omega$, and a logical connection between the frames given by $I$: we know that, for $\omega \in \Omega$, if $\omega$ is true, then $I(\omega)$ is also true. Here it is assumed that P is made with knowledge of $I$.

The reason for the last condition in the definition is that if $\omega$ is true then $I(\omega)$ is true; however, if $I(\omega) = \emptyset$ then, since $\emptyset$ is the contradictory proposition, $\omega$ cannot be true, so must be assigned zero probability.

Since it is frame $\Theta$ that we are interested in, we need to extend our uncertain information about $\Omega$ to $2^\Theta$. Associated with the source structure $S$ is a belief function and mass function over $\Theta$ (see [Shafer, 76a] for the definitions of these terms) defined, for $X \subseteq \Theta$ by

$$m^S(X) = \sum_{\omega \in \Omega : I(\omega) = X} P(\omega)$$

$$Bel^S(X) = \sum_{\omega \in \Omega : I(\omega) \subseteq X} P(\omega).$$

$Bel^S$ is the extension of the uncertain information given by P, via the compatibility function $I$, to the frame $\Theta$. It is viewed as a subjective measure of belief over $\Theta$, and is generally non-additive.

### 2.2 THE CONNECTION BETWEEN SOURCE STRUCTURES AND BELIEF FUNCTIONS

In his book, *a mathematical theory of of evidence* [Shafer, 76a], Shafer re-interprets Dempster's frame-work, and focuses on belief functions (the lower probabilities in Dempster's framework). The relationship between Dempster's and Shafer's frameworks is fairly straight-forward, but for clarity the connection will be described here. Although this paper deals primarily with source structures, and justifies Dempster's rule within Dempster's framework, these results also apply to Shafer's framework, using the correspondence between the two.

---
[3]See also 'Dempster spaces' in [Hájek et al., 92].

**Proposition**

Function Bel : $2^\Theta \rightarrow [0,1]$ is a belief function if and only if there exists a source structure $S$ over $\Theta$ with $Bel^S = Bel$.

Each belief function has a unique associated mass function, and vice versa. The focal elements of a belief function are the subsets of the frame which have non-zero mass. Let us define the focal elements of a source structure $S = (\Omega, P, I)$ over $\Theta$ to be the subsets $A$ of $\Theta$ such that $I(\omega) = A$ for some $\omega \in \Omega$ such that $P(\omega) \neq 0$. It can easily be seen that the set of focal elements of $S$ is the same as the set of focal elements of $Bel^S$.

From any belief function Bel, one can generate a source structure by letting $\Omega$ be a set in 1-1 correspondence with the set of focal elements, and defining the underlying probability function and compatibility function in the obvious way.

Though the underlying frame may be more abstract than the frame of interest, the natural occurrences of belief functions generally seem to have an intrinsic underlying frame. Even Shafer, who in his book does away with the underlying frame, uses a Dempster-type framework in later work, for example in his random codes justification of Dempster's rule.

### 2.3 EXTENSION TO DIFFERENT FRAMES OF INTEREST

It is assumed here that all the source structures we are interested in combining are over the same frame. This is not really a restriction since if they are over different frames, we can take a common refinement $\Theta$ of all the frames (see [Shafer, 76a, chapter 6]). All the source structures can then be re-expressed as source structures over $\Theta$, and we can proceed as before.

## 3  COMBINATION RULES

Crucial to Shafer's and Dempster's theories is combination of belief functions/source structures. The idea is that the body of evidence is broken up into small, (intuitively) independent pieces, the impact of each individual piece of evidence is represented by a belief function, and the impact of the whole body of evidence is calculated by combining these belief functions using Dempster's rule.

Informally, a combination rule is a mapping which takes a collection of source structures and gives a source structure, which is intended to represent the combined effect of the collection; the combined measures of belief in propositions of interest can then be calculated. If possible we would like to make natural assumptions that determine a uniquely sensible combination rule.

### 3.1 COMBINING SOURCE STRUCTURES

First a collection of source structures must be formally represented. This is done using a multiple source structure.

**Definition: Multiple Source Structures**

A multiple source structure $s$ over frame $\Theta$ is defined to be a function with finite domain $\psi^s \subset \mathbb{N}$, which maps each $i \in \psi^s$ to a source structure over $\Theta$; we write $s(i)$ as the triple $(\Omega_i^s, P_i^s, I_i^s)$.

There are some collections of source structures that give inconsistent information. This leads to the following definition, which is justified in section 6.

**Definition: Combinable**

Multiple source structure $s$ (over some frame) is said to be combinable if there exist $\omega_i \in \Omega_i^s$ (for each $i \in \psi^s$) with $P_i^s(\omega_i) \neq 0$ and $\bigcap_{i \in \psi^s} I_i^s(\omega_i) \neq \emptyset$.

**Definition: Combination Rule**

Let $\mathcal{C}$ be the set of all combinable multiple source structures (over any frames). A combination rule $\Pi$ is defined to be a function with domain $\mathcal{C}$ such that, for $s \in \mathcal{C}$ over frame $\Theta$, $\Pi(s)$ is a source structure over $\Theta$.

### 3.2 THE DIFFERENT COMPONENTS OF A COMBINATION RULE

It turns out that there are easy, natural choices for two of the three components of a combination rule, the two logical components.

**Definition**

For multiple source structure $s$,

(i) $\Omega^s$ is defined to be $\prod_{i \in \psi^s} \Omega_i^s$. An element $\omega$ of $\Omega^s$ is a function with domain $\psi^s$ such that $\omega(i) \in \Omega_i^s$. The element $\omega(i)$ will usually be written $\omega_i$.

(ii) The compatibility function $I^s$ is given by $I^s(\omega) = \bigcap_{i \in \psi^s} I_i^s(\omega_i)$.

**(i) The Underlying Frame**

Let us interpret element $\omega \in \Omega^s$ as meaning that $\omega_i$ is true for all $i \in \psi^s$. $\Omega^s$ is exhaustive, since each $\Omega_i^s$ is exhaustive, and every combination is considered; the elements of $\Omega^s$ are mutually exclusive since any two different $\omega$s differ in at least one co-ordinate $i$, and the elements of $\Omega_i^s$ are mutually exclusive. Therefore we can use $\Omega^s$ as the underlying frame for the combination. (Some of the elements of the product space may well be known to be impossible, using the compatibility functions, so a smaller underlying frame could be used, but this makes essentially no difference).

**(ii) The Combined Compatibility Function**

For $\omega \in \Omega^s$, if $\omega$ is true, then $\omega_i$ ($\in \Omega_i^s$) is true for each $i \in \psi^s$, which implies $I_i^s(\omega_i)$ is true for each $i$, so $\bigcap_{i \in \psi^s} I_i^s$ is true (since intersection of sets in $2^\Theta$ corresponds to conjunction of propositions). Assuming we have no other information about dependencies between underlying frames, this is the strongest proposition we can deduce from $\omega$. Thus compatibility functions $I_i^s$ generate compatibility function $I^s$ on $\Omega^s$.

**(iii) The Combined Underlying Probability Function**

This is the hard part of the combination rule so it is convenient to consider this part on its own, defining a C-rule to be the third component of a combination rule.

**Definition: C-rule**

A C-rule $\pi$ is defined to be a function, with domain the set of all combinable multiple source structures, which acts on a combinable multiple source structure $s$ over some frame $\Theta$ and produces an additive probability function over $\Omega^s$. We write the probability function $\pi(s)$ as $\pi^s$.

## 4 DEMPSTER'S RULE OF COMBINATION[4]

In this section Dempster's rule is expressed within the framework of this paper, and previous justifications are discussed.

---

[4]This refers to the rule described in [Shafer, 76a] and the combination rule in [Dempster, 67], not the amended non-normalised version of the rule, suggested in [Smets, 88], which is sometimes, confusingly, also referred to as 'Dempster's rule'.

## 4.1 DEMPSTER'S COMBINATION RULE AND C-RULE

**Definition: the Dempster C-rule**

The Dempster C-rule $\pi_{DS}$ is defined as follows. For combinable multiple source structure $s$, and $\omega \in \Omega^s$, if $I^s(\omega) = \emptyset$ then $\pi_{DS}^s(\omega) = 0$, else

$$\pi_{DS}^s(\omega) = K \prod_{i \in \psi^s} P_i^s(\omega_i),$$

where $K$ is a constant (i.e., independent of $\omega$) chosen such that $\pi_{DS}^s(\Omega^s) = 1$ (as it must for $\pi_{DS}^s$ to be a probability function).

**Definition: the Dempster Combination Rule**

The Dempster Combination Rule acts on multiple source structure $s$ to give source structure $(\Omega^s, \pi_{DS}^s, I^s)$.

It is easy to see that this is the combination rule used in [Dempster, 67] and corresponds to 'Dempster's rule' in [Shafer, 76a].

Justification of Dempster's rule therefore amounts to justifying the Dempster C-rule $\pi_{DS}$.

In section 6 the Dempster C-rule is justified by considering a set of constraints and assumptions on C-rules that determine a unique C-rule.

## 4.2 DISCUSSION OF JUSTIFICATIONS OF DEMPSTER'S RULE

Dempster's explanation of his rule in [Dempster, 67] amounts to assuming independence (so that for any $\omega \in \Omega^s$, the propositions represented by $\omega_i$ for $i \in \psi^s$ are considered to be independent) thus generating the product probability function $P(\omega) = \prod_{i \in \psi^s} P_i^s(\omega_i)$, for $\omega \in \Omega^s$. If $I^s(\omega)$ is empty then $\omega$ cannot be true, so P is then conditioned on the set $\{\omega : I^s(\omega) \neq \emptyset\}$, leading to Dempster's rule.

This two stage process, of firstly assuming independence, and then conditioning on $I^s(\omega)$ being non-empty, needs to be justified. The information given by $I^s$ is a dependence between $\omega_i$ for $i \in \psi^s$, so they clearly should not be assumed to be independent if this dependence is known. Ruspini's justification [Ruspini, 87] also appears not to deal satisfactorily with this crucial point.

A major weakness of *a mathematical theory of evidence* is that the numerical measures of belief are not given a clear interpretation, and Dempster's rule is not properly justified. This is rectified in [Shafer, 81] with his random codes canonical examples.

**Shafer's Random Codes Canonical Examples**

Here the underlying frame $\Omega$ is a set of codes. An agent randomly picks a particular code $\omega$ with chance $P(\omega)$ and this code is used to encode a true statement, which is represented by a subset of some frame $\Theta$. We know the set of codes and the chances of each being picked, but not the particular code picked, so when we receive the encoded message we decode it with each code $\omega' \in \Omega$ in turn to yield a message $I(\omega')$ (which is a subset of $\Theta$ for each $\omega'$). This situation corresponds to a source structure $(\Omega, P, I)$ over $\Theta$.

This leads to the desired two stage process: for if there are a number of agents picking codes stochastically independently and encoding true (but possibly different) messages then the probability distributions are (at this stage) independent. Then if we receive all their messages and decode them we may find certain combinations of codes are incompatible, leading to the second, conditioning, stage.

To use Shafer's theory to represent a piece of evidence, we choose the random codes canonical example (and associated source structure) that is most closely analogous to that piece of evidence. Two pieces of evidences are considered to be independent if we can satisfactorily compare them to the picking of independent random codes. However, in practice, it will often be very hard to say whether our evidences are analogous to random codes canonical examples, and judging whether these random codes are independent may also be very hard, especially if the comparison is a rather vague one.[5]

Shafer's justification applies only when the underlying probability function has meaning independently of the compatibility function, that is, when the compatibility function is *transitory* [Shafer, 92] (see also [Wilson, 92b] for some discussion of this point). Many occurrences of belief functions are not of this form. The justification given in this paper opens up the possibility of justifying Dempster's rule for other cases.

**The Non-Normalised Version of Dempster's Rule**

The non-normalised version of Dempster's rule [Smets, 88, 92] is simpler mathematically so it is less hard to find mathematical assumptions that determine it. However, whether these assumptions are reasonable or not is another matter. Smets considers that the unnormalised rule applies when the frame is interpreted as a set of mutually exclusive propositions which are not known to be exhaustive. Such a frame can be represented by a conventional frame, by adding an extra element representing the proposition which is true if and only if all the other propositions (represented by other elements of the frame) are false, thus restoring exhaustivity. Therefore Smets' non-exhaustive frames are unnecessary (and are restrictive).

---

[5] Other criticisms of this justification are given in the various comments on [Shafer, 82a, 82b], and in [Levi, 83].

Smets also attempts to justify (the normalised) Dempster's rule using the unnormalised rule by 'closed-world conditioning' [Smets, 88], i.e., combining the belief functions as if the frame was not known to be exhaustive, and then conditioning on the frame being exhaustive after all. This suffers from a similar problem to that faced by Dempster's justification (see above discussion), and seems very unsatisfactory: if we know that the frame is exhaustive then this information should be taken into account at the beginning (and then Smets' justification does not apply)—pretending temporarily that the frame is not exhaustive is perverse and liable to lead to unreliable results.

See also [Dubois and Prade, 86; Hájek, 92; Klawonn and Schwecke, 92].

## 5  BAYESIAN CONDITIONING

In this section Bayesian conditioning of source structures[6] is defined; these are used to simply express assumption (A) in section 6.

### Definition: Bayesian Conditioning of a Probability Function

Let P be an (additive) probability function over set $\Omega$, and let $\Delta \subseteq \Omega$ be such that $P(\Delta) \neq 0$. Then the probability function $P_\Delta$ over $\Delta$ is defined by

$$P_\Delta(\Gamma) = \frac{P(\Gamma)}{P(\Delta)} \quad \text{for } \Gamma \subseteq \Delta.$$

This is used for conditioning on certain evidence $\Delta$. Note that if $\Delta$ is considered to be certain, and $\Omega$ is a frame, then $\Delta$ is also a frame.

### Definition: Bayesian Conditioning of a Source Structure

Let $S = (\Omega, P, I)$ be a source structure over frame $\Theta$ and let $\Delta \subseteq \Omega$ (representing certain evidence) be such that $P(\Delta) \neq 0$. Then $S_\Delta$ is defined to be the source structure $(\Delta, P_\Delta, I_\Delta)$, where $I_\Delta$ is $I$ restricted to $\Delta$.

This should be uncontroversial, given that the judgement of the underlying epistemic probability P is made with knowledge of the compatibility function.

Incidentally if, for source structure $S = (\Omega, P, I)$ over $\Theta$ and $A \subseteq \Theta$, we let $\Delta = \{\omega \in \Omega : I(\omega) \subseteq A\}$ then $S_\Delta$ corresponds to geometric conditioning by $A$ [Shafer, 76b; Suppes and Zanotti, 77].

---

[6]This is not closely related to Bayesian updating of a belief function [Kyburg, 87; Jaffray, 92]

### Definition: Product Subsets

Let $s$ be a multiple source structure. $\Delta$ is said to be a product subset of $\Omega^s$ (with respect to $s$) if $\Delta = \prod_{i \in \psi^s} \Delta_i$ for some $\emptyset \neq \Delta_i \subseteq \Omega_i^s$ ($i \in \psi^s$).

Note that such a representation, if it exists, is unique. For product subset $\Delta$ of $\Omega^s$ and $i \in \psi^s$, we will write $\Delta_i$ as the projection of $\Delta$ into $\Omega_i^s$.

The following is a straight-forward extension of the Bayesian conditioning of a source structure.

### Definition: Bayesian Conditioning of a Multiple Source Structure

Let $s$ be a multiple source structure and let $\Delta$ be a product subset of $\Omega^s$ such that $P_i^s(\Delta_i) \neq 0$ for all $i \in \psi^s$. Then the multiple source structure $s_\Delta$ is defined as follows: $s_\Delta$ has domain $\psi^s$ and, for $i \in \psi^s$, $s_\Delta(i) = (s(i))_\Delta$.

## 6  CONSTRAINTS AND ASSUMPTIONS ON C-RULES

In this section we introduce two clearly natural constraints on C-rules, and two arguably reasonable assumptions. It is shown that together these determine a unique C-rule, which turns out to be Dempster's C-rule, hence justifying Dempster's rule.

### Constraint: Respecting Contradictions

C-rule $\pi$ is said to *respect contradictions* if for any combinable multiple source structure $s$ and $\omega \in \Omega^s$, if $I^s(\omega) = \emptyset$ then $\pi^s(\omega) = 0$.

If $I^s(\omega) = \emptyset$ then $\omega$ cannot be true since $\omega$ true implies $I^s(\omega)$ true, and $\emptyset$ represents the contradictory proposition. Therefore any sensible C-rule must respect contradictions.

### Constraint: Respecting Zero Probabilities

C-rule $\pi$ is said to *respect zero probabilities* if for any combinable multiple source structure $s$ and $\omega \in \Omega^s$, if $P_i^s(\omega_i) = 0$ for some $i \in \psi^s$, then $\pi^s(\omega) = 0$.

If $P_i^s(\omega_i) = 0$ for some $i$ then $\omega_i$ is considered impossible (since frames are finite), so, since $\omega$ is the conjunction of the propositions $\omega_i$, $\omega$ should clearly have zero probability.

Note that if we missed out the condition that the multiple source structure had to be 'combinable' in these two constraints and in the definition of a C-rule then these two constraints are inconsistent: for any C-rule $\pi$ and any multiple source structure $s$ which is not combinable, if $\pi$ respects contradictions and zero proba-bilities then $\pi^s(\omega) = 0$ for any $\omega \in \Omega^s$, which is inconsistent with $\pi^s$ being a probability function.

### Definition

Let $s$ be a multiple source structure, $k \in \psi^s$, and $l \in \Omega_k^s$. Then

$E_k^l$ is defined to be $\{\omega \in \Omega^s : \omega(k) = l\}$, and $\neg E_k^l$ is defined to be $\{\omega \in \Omega^s : \omega(k) \neq l\}$, i.e., $\Omega^s \setminus E_k^l$. The set $E_k^l$ is the cylindrical extension in $\Omega^s$ of $l$ ($\in \Omega_k^s$), and can be thought of as expressing the event that variable $k$ takes the value $l$.

### Definition: Assumption (A)

C-rule $\pi$ is said to satisfy assumption (A) if $\pi$ respects zero probabilities and, for any combinable multiple source structure $s$, for any $k \in \psi^s$, $l \in \Omega_k^s$ such that $\pi^s(\Delta) \neq 0$, where $\Delta = \neg E_k^l$,

$$(\pi^s)_\Delta = \pi^{s_\Delta}.$$

Note that since $\pi$ respects zero probabilities, if $\pi^s(\Delta) \neq 0$ then $P_k^s(\Delta_k) \neq 0$, so $s_\Delta$ is defined.

In fact it can be shown that if $\pi$ satisfies assumption (A) then it satisfies a more general form of the assumption where $\Delta$ is allowed to be an arbitrary product subset of $\Omega^s$.

Assumption (A) can be thought of as postulating that Bayesian conditioning commutes with source structure combination.

Bayesian conditioning by $\neg E_k^l$ can be viewed (roughly speaking) as omitting the $l^{\text{th}}$ focal element from the $k^{\text{th}}$ Belief function (and scaling up the other masses). Assumption (A) amounts to saying that it should not make any difference whether we omit that focal element before, or after, combination.

### Definition: Assumption (B)

Let $s$ be a combinable multiple source structure such that, for some $k \in \psi^s$,

$|\Omega_k^s| = 2$ and $|\Omega_i^s| = 1$ for $i \in \psi^s \setminus \{k\}$,

and $I^s(\omega) \neq \emptyset$ for $\omega \in \Omega^s$.

Then for $l \in \Omega_k^s$,

$$\pi^s(E_k^l) = P_k(l).$$

The notation hides the simplicity of this assumption. The multiple source structures referred to are of a very simple kind: one of the component source structures has just two elements in its underlying space, and so leads to a belief function with at most two focal elements, and all the other component source structures give belief functions with just one focal element, so they can be viewed as just propositions, i.e., certain evidences; furthermore there is no conflict in the evidences. In terms of belief functions this is the situation where we are conditioning a belief function with two masses by a subset of $\Theta$.

Assumption (B) is just that adding all the other certain sources does not change the probabilities of component $k$. The rationale behind this assumption is that the certain evidences are not in conflict with the information summarised by the $k^{\text{th}}$ source structure, so why should they change the probabilities?

### Theorem

$\pi_{DS}$ is the unique C-rule respecting contradictions, zero probabilities and satisfying (A) and (B).

This means that Dempster's rule of combination uniquely satisfies our constraints and assumptions, hence justifying it.

### Sketch of Proof

Unfortunately the proof of this theorem is far too long to be included here. To give the reader some idea of the structure of the proof, it will be briefly sketched.

It can easily be checked that $\pi_{DS}$ satisfies the constraints and assumptions. Conversely, let $\pi$ be an arbitrary C-rule satisfying the constraints and assumptions. First, it is shown that $\pi$ satisfies a more general form of (A), where $\Delta$ is allowed to be an arbitrary product subset of $\Omega^s$. This is then applied to the case of $\Delta = \{\omega, \omega'\}$ where $\omega$ and $\omega'$ differ in only one co-ordinate. In conjunction with assumption (B) this enables us to show that, when the denominators are non-zero,

$$\frac{\pi^s(\omega)}{\pi_{DS}^s(\omega)} = \frac{\pi^s(\omega')}{\pi_{DS}^s(\omega')}.$$

A source structure over $\Theta$ is said to be discounted if $\Theta$ is a focal element of it, and a multiple source structure is said to be discounted if each of the source structures of which it is composed is discounted. It is then shown that, for any discounted multiple source structure $s$, $\pi^s = \pi_{DS}^s$, using the last result repeatedly. The theorem is then proved by taking an arbitrary combinable multiple source structure $t$, discounting it to form $s$ (see [Shafer, 76a]) and using the more general form of assumption (A) again to relate $\pi^t$ and $\pi^s = \pi_{DS}^s$.

## 7 DISCUSSION

Both assumptions (A) and (B) seem fairly reasonable. (A) appears to be an attractive property of a C-rule, but is a rather strong one, and it is not currently clear to me in which situations it should hold (it is conceivable that there are other reasonable-seeming principles with which it is sometimes in conflict). Further work should attempt to clarify exactly when both assumptions are reasonable.

There are cases where Dempster's rule can seem unintuitive, for example, I argued in [Wilson, 92b] that Dempster's rule is unreasonable at least for some instances of Bayesian belief functions, and there has been much criticism of certain examples of the use of the rule e.g., [Pearl, 90a, 90b; Walley, 91; Voorbraak, 91; Zadeh, 84].[7]

If it does turn out that there are certain types of belief functions where assumption (A) or (B) is not reasonable, then the above theorem, as it stands, is not useful. However, an examination of its proof reveals that only two operations on belief functions/source structures are used—Bayesian conditioning (i.e, omitting focal elements and scaling the others up) and discounting (i.e, adding a focal element equal to the frame $\Theta$, and scaling the others down). This means that the proof could be used to justify Dempster's rule for any sub-class of belief functions/source structures (for which (A) and (B) may be more reasonable) which is closed under these operations, for example the set of simple support functions or the set of consonant support functions. Also, for the same reason, the proof could be used to justify Dempster's rule for collections of belief functions/multiple source structures $s$ such that $I^s(\omega) \neq \emptyset$ for all $\omega \in \Omega^s$, if (A) and (B) were considered reasonable here.

It might also be interesting to investigate alternatives to (B), which give different values for $\pi^s(E_k^l)$ than those given in (B). The proof of the theorem can be modified to show that there is at most one C-rule satisfying the constraints and assumptions, though of course it will not be the Dempster C-rule.

## Acknowledgements

I am very grateful to an anonymous referee for pointing out a minor error.

This work was supported by a SERC postdoctoral fellowship, based at Queen Mary and Westfield College. Thanks also to Oxford Brookes University for use of their facilities.

---

[7]Most of which, though, I find unconvincing; for example, see [Wilson, 92a] and the other papers in the same issue for discussion of Pearl's criticisms; Zadeh's criticisms were convincingly answered in [Shafer, 84].